\documentclass[11pt,a4paper]{article}
\usepackage[hyperref]{naaclhlt2018}
\usepackage{times}
\usepackage{latexsym}
\usepackage{ctable}
\usepackage{subcaption}
\usepackage{color}
\usepackage[justification=centering]{caption}
\usepackage{booktabs, siunitx}
\usepackage{graphicx}
\usepackage[export]{adjustbox}
\usepackage{placeins}
\usepackage{cprotect}
\usepackage{moresize}
\usepackage[font=small, labelfont=bf, format=plain, justification=justified, singlelinecheck=false]{caption}
\usepackage{soul}

\usepackage{url}
\aclfinalcopy 
\def\aclpaperid{783} 

\title{When and Why are Pre-trained Word Embeddings Useful \\
       for Neural Machine Translation?}

\author{\fontsize{11.5}{12} \selectfont Ye Qi, Devendra Singh Sachan, Matthieu Felix, Sarguna Janani Padmanabhan, Graham Neubig \\
  Language Technologies Institute, Carnegie Mellon University \\
  {\tt \{yeq,dsachan,matthief,sjpadman\}@andrew.cmu.edu, gneubig@cs.cmu.edu}}

\date{}

\begin{document}
\maketitle

\begin{abstract}
The performance of Neural Machine Translation (NMT) systems often suffers in low-resource scenarios where sufficiently large-scale parallel corpora cannot be obtained. Pre-trained word embeddings have proven to be invaluable for improving performance in natural language analysis tasks, which often suffer from paucity of data. However, their utility for NMT has not been extensively explored. In this work, we perform five sets of experiments that analyze when we can expect pre-trained word embeddings to help in NMT tasks. We show that such embeddings can be surprisingly effective in some cases -- providing gains of up to 20 BLEU points in the most favorable setting.%
\footnote{Scripts/data to replicate experiments are available at \url{https://github.com/neulab/word-embeddings-for-nmt}}
\end{abstract}

\section{Introduction}

Pre-trained word embeddings have proven to be highly useful in neural network models for NLP tasks such as sequence tagging \cite{Lample2016NeuralAF, ma-hovy:2016:P16-1} and text classification  \cite{kim2014convolutional}. However, it is much less common to use such pre-training in NMT ~\cite{Wu2016GooglesNM}, largely because the large-scale training corpora used for tasks such as WMT\footnote{\url{http://www.statmt.org/wmt17/}} tend to be several orders of magnitude larger than the annotated data available for other tasks, such as the Penn Treebank \cite{marcus1993building}. However, for low-resource languages or domains, it is not necessarily the case that bilingual data is available in abundance, and therefore the effective use of monolingual data becomes a more desirable option.

Researchers have worked on a number of methods for using monolingual data in NMT systems \cite{cheng2016semi,he2016dual,ramachandran2016unsupervised}.
Among these, pre-trained word embeddings have been used either in standard translation systems \cite{neishi-EtAl:2017:WAT2017,artetxe2017unsupervised} or as a method for learning translation lexicons in an entirely unsupervised manner \cite{conneau2017word, digangi17lowresource}. Both methods show potential improvements in BLEU score when pre-training is properly integrated into the NMT system.

However, from these works, it is still not clear as to \emph{when} we can expect pre-trained embeddings to be useful in NMT, or \emph{why} they provide performance improvements. In this paper, we examine these questions more closely, conducting five sets of experiments to answer the following questions:

\begin{itemize}
\setlength\itemsep{0mm}
\item[Q1] Is the behavior of pre-training affected by language families and other linguistic features of source and target languages? (\S\ref{sec:efficacy})
\item[Q2] Do pre-trained embeddings help more when the size of the training data is small? (\S\ref{sec:data-size})
\item[Q3] How much does the similarity of the source and target languages affect the efficacy of using pre-trained embeddings? (\S\ref{sec:similarity})
\item[Q4] Is it helpful to align the embedding spaces between the source and target languages? (\S\ref{sec:alignment})
\item[Q5] Do pre-trained embeddings help more in multilingual systems as compared to bilingual systems? (\S\ref{sec:multilinguality})
\end{itemize}

\section{Experimental Setup}

In order to perform experiments in a controlled, multilingual setting, we created a parallel corpus from TED talks transcripts.\footnote{\url{https://www.ted.com/participate/translate}}
Specifically, we prepare data between English (\textsc{En}) and three pairs of languages, where the two languages in the pair are similar, with one being relatively low-resourced compared to the other: Galician (\textsc{Gl}) and Portuguese (\textsc{Pt}), Azerbaijani (\textsc{Az}) and Turkish (\textsc{Tr}), and Belarusian (\textsc{Be}) and Russian (\textsc{Ru}). The languages in each pair are similar in vocabulary, grammar and sentence structure ~\cite{matthews1997concise}, which controls for language characteristics and also improves the possibility of transfer learning in multi-lingual models (in \S\ref{sec:multilinguality}).
They also represent different language families -- \textsc{Gl}/\textsc{Pt} are Romance; \textsc{Az}/\textsc{Tr} are Turkic; \textsc{Be}/\textsc{Ru} are Slavic -- allowing for comparison across languages with different caracteristics.
Tokenization was done using \verb+Moses+ tokenizer\footnote{\url{https://github.com/moses-smt/mosesdecoder/blob/master/scripts/tokenizer/tokenizer.perl}} and hard punctuation symbols were used to identify sentence boundaries.
Table~\ref{tab:dataset-size} shows data sizes.

\begin{table}[t!]
\small
\begin{center}
\begin{tabular}{r|rrr}
  \toprule
  \bf Dataset & \tt train & \tt dev & \tt test \\	
  \midrule
  \textsc{Gl} $\rightarrow$ \textsc{En} & $10,017$ & $682$ & $1,007$\\
  \textsc{Pt} $\rightarrow$ \textsc{En} & $51,785$ & $1,193$ & $1,803$\\
  \textsc{Az} $\rightarrow$ \textsc{En} & $5,946$ & $671$ & $903$\\
  \textsc{Tr} $\rightarrow$ \textsc{En} & $182,450$ & $4,045$ & $5,029$\\
  \textsc{Be} $\rightarrow$ \textsc{En} & $4,509$ & $248$ & $664$\\
  \textsc{Ru} $\rightarrow$ \textsc{En} & $208,106$ & $4,805$ & $5,476$\\
  \bottomrule
\end{tabular}
\vspace{-2mm}
\caption{\label{tab:dataset-size}Number of sentences for each language pair.}
\vspace{-4mm}
\end{center}
\end{table}



For our experiments, we use a standard 1-layer encoder-decoder model with attention \cite{bahdanau+al-2014-nmt} with a beam size of $5$ implemented in \verb+xnmt+\footnote{\url{https://github.com/neulab/xnmt/}} \cite{neubig18xnmt}. Training uses a batch size of $32$ and the Adam optimizer \cite{kingma2014adam} with an initial learning rate of $0.0002$, decaying the learning rate by $0.5$ when development loss decreases \cite{denkowski2017stronger}. We evaluate the model's performance using BLEU metric \cite{papineni2002bleu}.

We use available pre-trained word embeddings \cite{bojanowski2016enriching} trained using \verb+fastText+\footnote{\url{https://github.com/facebookresearch/fastText/}} on Wikipedia\footnote{\url{https://dumps.wikimedia.org/}} for each language. These word embeddings \cite{mikolov2017wordvectors} incorporate character-level, phrase-level and positional information of words and are trained using CBOW algorithm \cite{mikolov2013distributed}. The dimension of word embeddings is set to $300$. The embedding layer weights of our model are initialized using these pre-trained word vectors. In baseline models without pre-training, we use \newcite{glorot2010understanding}'s uniform initialization.

\section{Q1: Efficacy of Pre-training}
\label{sec:efficacy}
In our first set of experiments, we examine the efficacy of pre-trained word embeddings across the various languages in our corpus.
In addition to providing additional experimental evidence supporting the findings of other recent work on using pre-trained embeddings in NMT \cite{neishi-EtAl:2017:WAT2017,artetxe2017unsupervised,digangi17lowresource}, we also examine whether pre-training is useful across a wider variety of language pairs and if it is more useful on the source or target side of a translation pair.

\begin{table}[t!]
\small
\begin{center}
\begin{tabular}{r|rrrr}
  \toprule
  \bf Src $\rightarrow$ \hspace{0.5cm} & \tt std & \tt pre & \tt std & \tt pre \\	
  $\rightarrow$ \bf Trg & \tt std & \tt std & \tt pre & \tt pre \\	
  \midrule
  \textsc{Gl} $\rightarrow$ \textsc{En} & $2.2$ & $\mathbf{13.2}$ & $2.8$ & $12.8$ \\
  \textsc{Pt} $\rightarrow$ \textsc{En} & $26.2$ & $\mathbf{30.3}$ & $26.1$ & $\mathbf{30.8}$ \\
  \midrule
  \textsc{Az} $\rightarrow$ \textsc{En} & $1.3$ & $\mathbf{2.0}$ & $1.6$ & $\mathbf{2.0}$ \\
  \textsc{Tr} $\rightarrow$ \textsc{En} & $14.9$ & $17.6$ & $14.7$ & $\mathbf{17.9}$ \\
  \midrule
  \textsc{Be} $\rightarrow$ \textsc{En} & $1.6$ & $2.5$ & $1.3$ & $\mathbf{3.0}$ \\
  \textsc{Ru} $\rightarrow$ \textsc{En} & $18.5$ & $\mathbf{21.2}$ & $18.7$ & $\mathbf{21.1}$ \\
  \bottomrule
\end{tabular}
\end{center}
\vspace{-2mm}
\caption{\label{tab:q1}Effect of pre-training on BLEU score over six languages. The systems use either random initialization (\texttt{std}) or pre-training (\texttt{pre}) on both the source and target sides. }
\vspace{-4mm}
\end{table}

The results in Table~\ref{tab:q1} clearly demonstrate that pre-training the word embeddings in the source and/or target languages helps to increase the BLEU scores to some degree.
Comparing the second and third columns, we can see the increase is much more significant with pre-trained source language embeddings.
This indicates that the majority of the gain from pre-trained word embeddings results from a better encoding of the source sentence. 

The gains from pre-training in the higher-resource languages are consistent: $\approx$3 BLEU points for all three language pairs.
In contrast, for the extremely low-resource languages, the gains are either quite small (\textsc{Az} and \textsc{Be}) or very large, as in \textsc{Gl} which achieves a gain of up to 11 BLEU points. This finding is interesting in that it indicates that word embeddings may be particularly useful to bootstrap models that are on the threshold of being able to produce reasonable translations, as is the case for \textsc{Gl} in our experiments.

\section{Q2: Effect of Training Data Size}
\label{sec:data-size}

The previous experiment had interesting implications regarding available data size and effect of pre-training.
Our next series of experiments examines this effect in a more controlled environment by down-sampling the training data for the higher-resource languages to 1/2, 1/4 and 1/8 of their original sizes.

\begin{figure}[t]
\centering
\begin{subfigure}{.45\textwidth}
  \small
  \includegraphics[width=\linewidth]{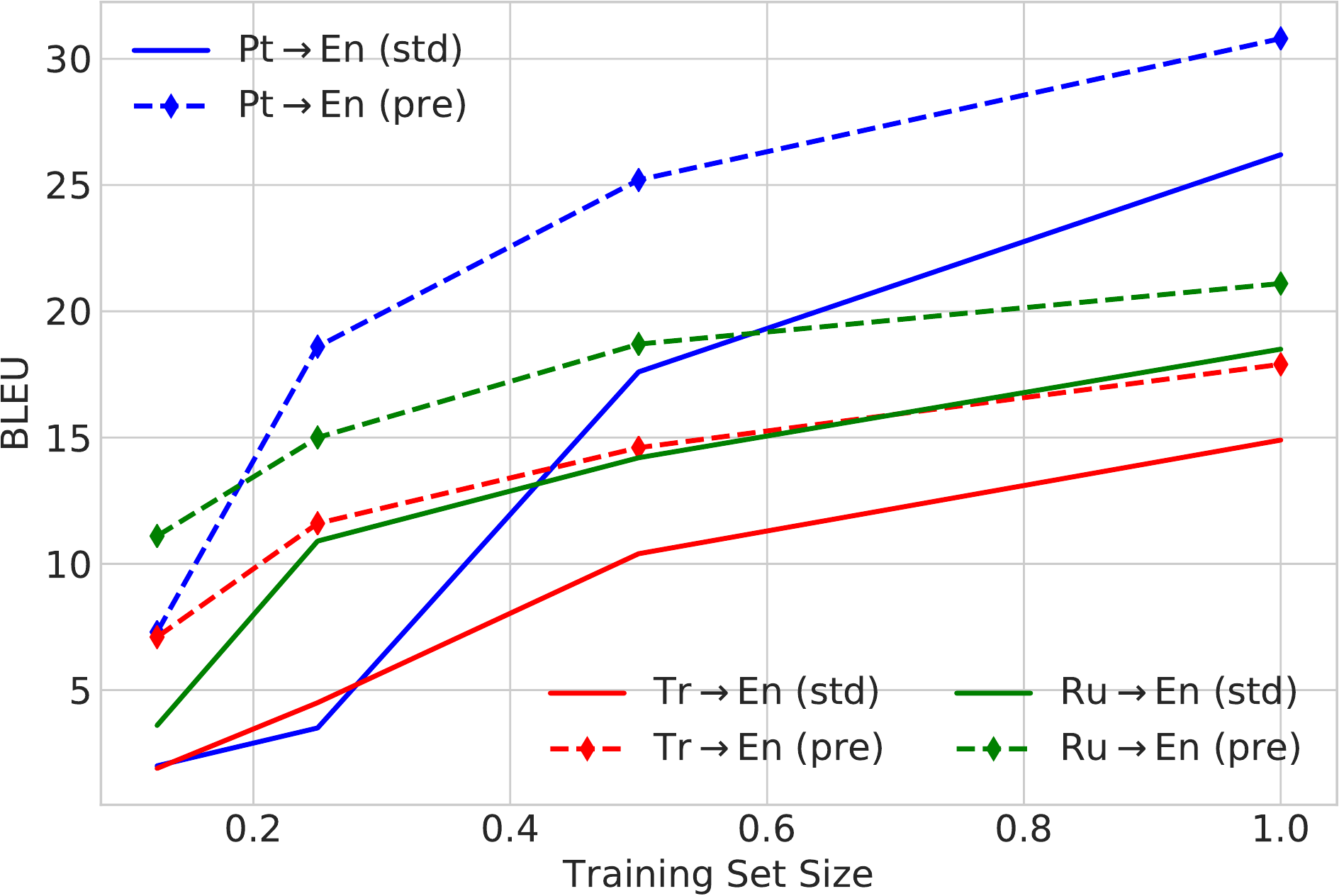}
  \vspace{-6mm}
\end{subfigure} \\
\begin{subfigure}{.45\textwidth}
  \small
  \includegraphics[width=\linewidth]{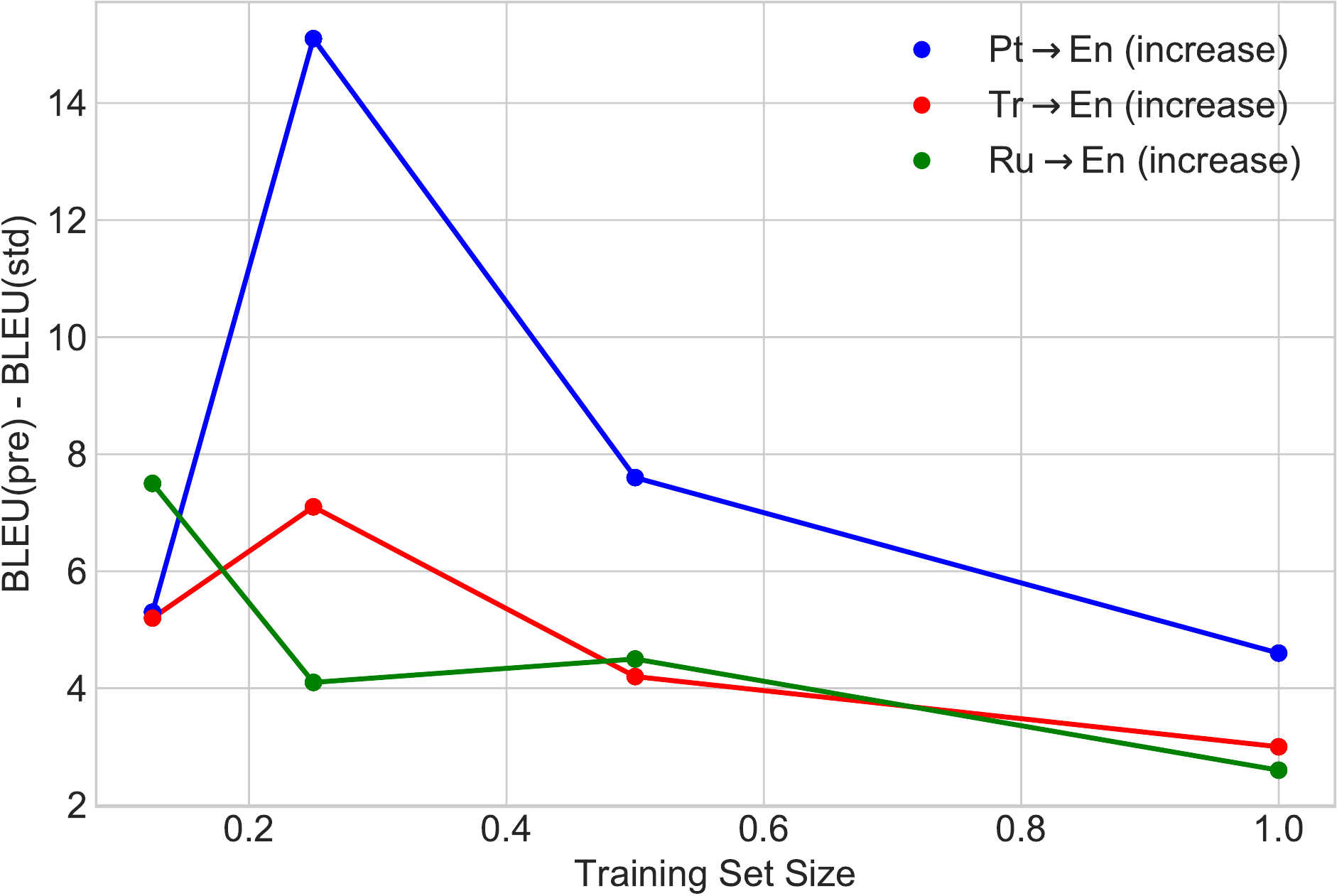}
  \vspace{-6mm}
\end{subfigure}
\centering
\caption{\label{fig:q2}BLEU and BLEU gain by data size. }
\vspace{-4mm}
\end{figure}


From the BLEU scores in Figure~\ref{fig:q2}, we can see that for all three languages the gain in BLEU score demonstrates a similar trend to that found in $\textsc{Gl}$ in the previous section: the gain is highest when the baseline system is poor but not too poor, usually with a baseline BLEU score in the range of 3-4. 
This suggests that at least a moderately effective system is necessary before pre-training takes effect, but once there is enough data to capture the basic characteristics of the language, pre-training can be highly effective.

\section{Q3: Effect of Language Similarity}
\label{sec:similarity}

The main intuitive hypothesis as to why pre-training works is that the embedding space becomes more consistent, with semantically similar words closer together.
We can also make an additional hypothesis: if the two languages in the translation pair are more linguistically similar, the semantic neighborhoods will be more similar between the two languages (i.e. semantic distinctions or polysemy will likely manifest themselves in more similar ways across more similar languages).
As a result, we may expect that the gain from pre-training of embeddings may be larger when the source and target languages are more similar. To examine this hypothesis, we selected Portuguese as the target language, which when following its language family tree from top to bottom, belongs to Indo-European, Romance, Western Romance, and West-Iberian families. We then selected one source language from each family above.\footnote{English was excluded because the TED talks were originally in English, which results in it having much higher BLEU scores than the other languages due to it being direct translation instead of pivoted through English like the others.} To avoid the effects of training set size, all pairs were trained on 40,000 sentences.

\begin{table}[t!]
\small
\begin{center}
\begin{tabular}{r|c|rr}
  \toprule
  \bf Dataset & \bf Lang. Family & \tt std & \tt pre \\	
  \midrule
  \textsc{Es} $\rightarrow$ \textsc{Pt} & West-Iberian & $17.8$ & $24.8$ $(+\mathbf{7.0})$ \\
  \textsc{Fr} $\rightarrow$ \textsc{Pt} & Western Romance & $12.4$ & $18.1$ $(+5.7)$ \\
  \textsc{It} $\rightarrow$ \textsc{Pt} & Romance & $14.5$ & $19.2$ $(+4.7)$ \\
  \textsc{Ru} $\rightarrow$ \textsc{Pt} & Indo-European & $2.4$ & $8.6$ $(+6.2)$ \\
  \textsc{He} $\rightarrow$ \textsc{Pt} & \emph{No Common} & $3.0$ & $11.9$ $(+\mathbf{8.9})$ \\
  \bottomrule
\end{tabular}
\end{center}
\vspace{-4mm}
\caption{\label{tab:q3} Effect of linguistic similarity and pre-training on BLEU. The language family in the second column is the most recent common ancestor of source and target language. }
\vspace{-4mm}
\end{table}

From Table~\ref{tab:q3}, we can see that the BLEU scores of \textsc{Es}, \textsc{Fr}, and \textsc{It} do generally follow this hypothesis. 
As we move to very different languages, \textsc{Ru} and \textsc{He} see larger accuracy gains than their more similar counterparts \textsc{Fr} and \textsc{It}.
This can be largely attributed to the observation from the previous section that systems with larger headroom to improve tend to see larger increases; \textsc{Ru} and \textsc{He} have very low baseline BLEU scores, so it makes sense that their increases would be larger.

\section{Q4: Effect of Word Embedding Alignment}
\label{sec:alignment}

Until now, we have been using embeddings that have been trained independently in the source and target languages, and as a result there will not necessarily be a direct correspondence between the embedding spaces in both languages.
However, we can postulate that having consistent embedding spaces across the two languages may be beneficial, as it would allow the NMT system to more easily learn correspondences between the source and target.
To test this hypothesis, we adopted the approach proposed by ~\newcite{smith2017offline} to learn orthogonal transformations that convert the word embeddings of multiple languages to a single space and used these aligned embeddings instead of independent ones.

\begin{table}[t!]
\small
\begin{center}
\begin{tabular}{r|rr}
  \toprule
  \bf Dataset & \tt unaligned & \tt aligned \\	
  \midrule
  \textsc{Gl} $\rightarrow$ \textsc{En} & $12.8$ & $11.5$ $(-1.3)$ \\
  \textsc{Pt} $\rightarrow$ \textsc{En} & $30.8$ & $30.6$ $(-0.2)$ \\
  \midrule
  \textsc{Az} $\rightarrow$ \textsc{En} & $2.0$ & 2.1 $(+0.1)$ \\
  \textsc{Tr} $\rightarrow$ \textsc{En} & $17.9$ & 17.7 $(-0.2)$ \\
  \midrule
  \textsc{Be} $\rightarrow$ \textsc{En} & $3.0$ & $3.0$ $(+0.0)$ \\
  \textsc{Ru} $\rightarrow$ \textsc{En} & $21.1$ & $21.4$ $(+0.3)$ \\
  \bottomrule
\end{tabular}
\end{center}
\vspace{-4mm}
\caption{\label{tab:q4}Correlation between word embedding alignment and BLEU score in bilingual translation task. }
\vspace{-4mm}
\end{table}

From Table~\ref{tab:q4}, we can see that somewhat surprisingly, the alignment of word embeddings was not beneficial for training, with gains or losses essentially being insignificant across all languages.
This, in a way, is good news, as it indicates that \emph{a priori} alignment of embeddings may not be necessary in the context of NMT, since the NMT system can already learn a reasonable projection of word embeddings during its normal training process.

\section{Q5: Effect of Multilinguality}
\label{sec:multilinguality}

Finally, it is of interest to consider pre-training in multilingual translation systems that share an encoder or decoder between multiple languages \cite{johnson16multilingual, firat2016multi}, which is another promising way to use additional data (this time from another language) as a way to improve NMT.
Specifically, we train a model using our pairs of similar low-resource and higher-resource languages, and test on only the low-resource language. For those three pairs, the similarity of \textsc{Gl}/\textsc{Pt} is the highest while \textsc{Be}/\textsc{Ru} is the lowest. 

\begin{table}[t!]
\small
\begin{center}
\begin{tabular}{cc|r|rrr}
  \toprule
  \bf Train & \bf Eval & \tt bi & \tt std & \tt pre & \tt align \\	
  \midrule
  \textsc{Gl} $+$ \textsc{Pt} & \textsc{Gl} & $2.2$ & $17.5$ & $20.8$ & $\mathbf{22.4}$ \\
  \textsc{Az} $+$ \textsc{Tr} & \textsc{Az} & $1.3$ & $5.4$ & $5.9$ & $\mathbf{7.5}$ \\
  \textsc{Be} $+$ \textsc{Ru} & \textsc{Be} & $1.6$ & $\mathbf{10.0}$ & $7.9$ & $9.6$ \\
  \bottomrule
\end{tabular}
\end{center}
\vspace{-4mm}
\caption{\label{tab:q5}Effect of pre-training on multilingual translation into English. \texttt{bi} is a bilingual system trained on only the eval source language and all others are multi-lingual systems trained on two similar source languages.}
\vspace{-4mm}
\end{table}

We report the results in Table~\ref{tab:q5}.
When applying pre-trained embeddings, the gains in each translation pair are roughly in order of their similarity, with \textsc{Gl}/\textsc{Pt} showing the largest gains, and \textsc{Be}/\textsc{Ru} showing a small decrease.
In addition, it is also interesting to note that as opposed to previous section, aligning the word embeddings helps to increase the BLEU scores for all three tasks.
These increases are intuitive, as a single encoder is used for both of the source languages, and the encoder would have to learn a significantly more complicated transform of the input if the word embeddings for the languages were in a semantically separate space.
Pre-training and alignment ensures that the word embeddings of the two source languages are put into similar vector spaces, allowing the model to learn in a similar fashion as it would if training on a single language.

Interestingly, \textsc{Be} $\rightarrow$ \textsc{En} does not seem to benefit from pre-training in the multilingual scenario, which hypothesize is due to the fact that: 1) Belarusian and Russian are only partially mutually intelligible ~\cite{corbett2003slavonic}, in another words, they are not as similar; 2) the Slavic languages have comparatively rich morphology, making sparsity in the trained embeddings a larger problem. 

\section{Analysis}

\subsection{Qualitative Analysis}

\begin{table*}[!t]
\begin{center}
\small
\begin{tabular}{l|p{12.5cm}}
  \toprule
  \tt source & \textit{( risos )} e \'e que \textbf{chris} \'e \ul{un grande avogado} , pero non sab\'ia case nada sobre \ul{lexislaci\'on de patentes} e absolutamente nada sobre xen\'etica .\\
  \tt reference & \textit{( laughter )} now \textbf{chris} is \ul{a really brilliant lawyer} , but he knew almost nothing about \ul{patent law} and certainly nothing about genetics . \\
  \tt bi:std & \textit{( laughter )} and \textbf{i} 'm not \ul{a little bit of a little bit of a little bit of} and ( laughter ) and i 'm going to be able to be a lot of years . \\
  \tt multi:pre-align & \textit{( laughter )} and \textbf{chris} is \ul{a big lawyer} , but i did n't know almost anything about \ul{patent legislation} and absolutely nothing about genetic . \\
  \bottomrule
\end{tabular}
\vspace{-3mm}
\caption{ \label{tab:exp-translation} Example translations of \textsc{Gl} $\rightarrow$ \textsc{En}.}
\vspace{-4mm}
\end{center}
\end{table*}

Finally, we perform a qualitative analysis of the translations from \textsc{Gl} $\rightarrow$ \textsc{En}, which showed one of the largest increases in quantitative numbers.
As can be seen from Table \ref{tab:exp-translation}, pre-training not only helps the model to capture rarer vocabulary but also generates sentences that are more grammatically well-formed. As highlighted in the table cells, the best system successfully translates a person's name (``\textit{chris}'') and two multi-word phrases (``\textit{big lawyer}'' and ``\textit{patent legislation}''), indicating the usefulness of pre-trained embeddings in providing a better representations of less frequent concepts when used with low-resource languages.

In contrast, the bilingual model without pre-trained embeddings substitutes these phrases for common ones (``\textit{i}''), drops them entirely, or produces grammatically incorrect sentences. The incomprehension of core vocabulary causes deviation of the sentence semantics and thus increases the uncertainty in predicting next words, generating several phrasal loops which are typical in NMT systems.

More qualitative examples and analysis are presented in the supplementary materials.

\subsection{Analysis of Frequently Generated $n$-grams.}

\begin{table*}[!htb]
\small
\begin{center}
\begin{subtable}{.45\textwidth}
\begin{center}
\begin{tabular}{lr|lr}
  \toprule
  \multicolumn{2}{c}{\tt bi:std} & \multicolumn{2}{|c}{\tt bi:pre} \\ 
  \midrule
  \tt ) so & $2/0$ & \tt about & $0/53$ \\
  \tt ( laughter ) i & $2/0$ & \tt people & $0/49$ \\
  \tt ) i & $2/0$ & \tt or & $0/43$ \\
  \tt laughter ) i & $2/0$ & \tt these & $0/39$ \\
  \tt ) and & $2/0$ & \tt with & $0/38$ \\
  \tt they were & $1/0$ & \tt because & $0/37$ \\
  \tt have to & $5/2$ & \tt like & $0/36$ \\
  \tt a new & $1/0$ & \tt could & $0/35$ \\
  \tt to do , & $1/0$ & \tt all & $0/34$ \\
  \tt `` and then & $1/0$ & \tt two & $0/32$ \\
  \bottomrule
\end{tabular}
\caption{Pairwise comparison between two bilingual models}
\end{center}
\end{subtable}
\begin{subtable}{.54\textwidth}
\begin{center}
\begin{tabular}{lr|lr}
  \toprule
  \multicolumn{2}{c}{\tt multi:std} & \multicolumn{2}{|c}{\tt multi:pre+align} \\ 
  \midrule
  \tt here & $6/0$ & \tt on the & $0/14$ \\
  \tt again , & $4/0$ & \tt like & $1/20$ \\
  \tt several & $4/0$ & \tt should & $0/ 9$ \\
  \tt you 're going & $4/0$ & \tt court & $0/ 9$ \\
  \tt 've & $4/0$ & \tt judge & $0/ 7$ \\
  \tt we 've & $4/0$ & \tt testosterone & $0/ 6$ \\
  \tt you 're going to & $4/0$ & \tt patents & $0/ 6$ \\
  \tt people , & $4/0$ & \tt patent & $0/ 6$ \\
  \tt what are & $3/0$ & \tt test & $0/ 6$ \\
  \tt the room & $3/0$ & \tt with & $1/12$ \\
  \bottomrule
\end{tabular}
\caption{Pairwise comparison between two multilingual models }
\end{center}
\end{subtable}
\caption{\label{tab:n-gram}Top 10 n-grams that one system did a better job of producing. The numbers in the figure, separated by a slash, indicate how many times each n-gram is generated by each of the two systems.}
\end{center}
\end{table*}

We additionally performed pairwise comparisons between the top 10 n-grams that each system (selected from the task \textsc{Gl} $\rightarrow$ \textsc{En}) is better at generating, to further understand what kind of words pre-training is particularly helpful for.\footnote{Analysis was performed using \texttt{compare-mt.py} from \url{https://github.com/neubig/util-scripts/}.}
The results displayed in Table \ref{tab:n-gram} demonstrate that pre-training helps both with words of low frequency in the training corpus, and even with function words such as prepositions.
On the other hand, the improvements in systems without pre-trained embeddings were not very consistent, and largely focused on high-frequency words.

\subsection{F-measure of Target Words}

\begin{figure}[!htb]
\centering
  \centering
  \includegraphics[width=\linewidth]{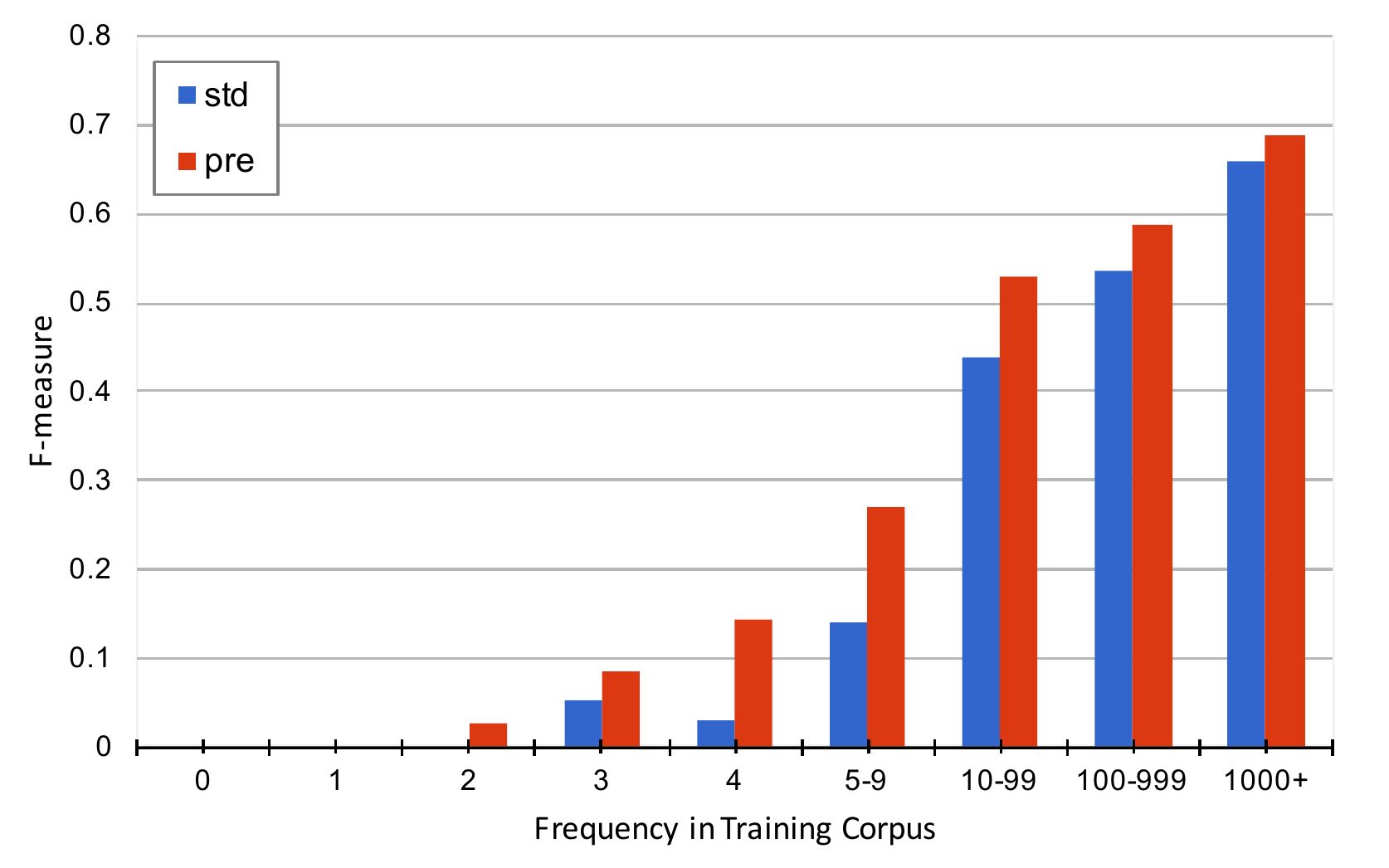}
  \centering
  \caption{\label{fig:fmeas} The f-measure of target words in bilingual translation task  \textsc{Pt} $\rightarrow$ \textsc{En}}
\end{figure}

Finally, we performed a comparison of the f-measure of target words, bucketed by frequency in the training corpus.
As displayed in Figure \ref{fig:fmeas}, this shows that pre-training manages to improve the accuracy of translation for the entire vocabulary, but particularly for words that are of low frequency in the training corpus.

\section{Conclusion}

This paper examined the utility of considering pre-trained word embeddings in NMT from a number of angles.
Our conclusions have practical effects on the recommendations for when and why pre-trained embeddings may be effective in NMT, particularly in low-resource scenarios:
(1) there is a sweet-spot where word embeddings are most effective, where there is very little training data but not so little that the system cannot be trained at all,
(2) pre-trained embeddings seem to be more effective for more similar translation pairs,
(3) \emph{a priori} alignment of embeddings may not be necessary in bilingual scenarios, but is helpful in multi-lingual training scenarios.

\section*{Acknowledgements}

Parts of this work were sponsored by Defense Advanced Research Projects Agency Information Innovation Office (I2O). Program: Low Resource Languages for Emergent Incidents (LORELEI). Issued by DARPA/I2O under Contract No. HR0011-15-C-0114. The views and conclusions contained in this document are those of the authors and should not be interpreted as representing the official policies, either expressed or implied, of the U.S. Government. The U.S. Government is authorized to reproduce and distribute reprints for Government purposes notwithstanding any copyright notation here on.

\bibliography{naaclhlt2018}
\bibliographystyle{acl_natbib}


\end{document}